\documentclass[3p,12pt, authoryear,english]{elsarticle}

\usepackage[english]{babel}
\usepackage[utf8]{inputenc}
\usepackage{graphicx}
\usepackage{url}
\usepackage{datetime}
\usepackage{lineno}
\usepackage{amsmath}
\usepackage{todonotes}
\usepackage{booktabs}
\usepackage{algorithm} 
\usepackage{algpseudocode}
\usepackage{breakcites}

\input amssym.tex
\begin{document}

\begin{frontmatter}

\title{Forecasting Across Time Series Databases using Recurrent Neural Networks on Groups of Similar Series: A Clustering Approach}

\author[monash1]{Kasun Bandara}
\author[monash1]{Christoph~Bergmeir\corref{cor1}}
\author[uber]{Slawek~Smyl}

\address[monash1]{Faculty of Information Technology, Monash University, Melbourne, Australia.}
\address[uber]{Uber Technologies Inc, San Francisco, California, United States.}

\cortext[cor1]{Corresponding author. Postal Address: Faculty of Information Technology, P.O. Box 63 Monash University, Victoria 3800, Australia. E-mail address: christoph.bergmeir@monash.edu Tel: +61 3 990 59555}

\begin{abstract}
With the advent of Big Data, nowadays in many applications databases containing large quantities of similar time series are available. Forecasting time series in these domains with traditional univariate forecasting procedures leaves great potentials for producing accurate forecasts untapped. Recurrent neural networks (RNNs), and in particular Long Short Term Memory (LSTM) networks, have proven recently that they are able to outperform state-of-the-art univariate time series forecasting methods in this context, when trained across all available time series. However, if the time series database is heterogeneous, accuracy may degenerate, so that on the way towards fully automatic forecasting methods in this space, a notion of similarity between the time series needs to be built into the methods. To this end, we present a prediction model that can be used with different types of RNN models on subgroups of similar time series, which are identified by time series clustering techniques. We assess our proposed methodology using LSTM networks, a widely popular RNN variant. Our method achieves competitive results on benchmarking datasets under competition evaluation procedures. In particular, in terms of mean sMAPE accuracy it consistently outperforms the baseline LSTM model, and outperforms all other methods on the CIF2016 forecasting competition dataset.

%
%
%
%
\end{abstract}

\end{frontmatter}


\section{Introduction}
\label{sec:intro}

In the time series forecasting community there has been the long-standing consensus that sophisticated methods do not necessarily produce better forecasts than simpler ones. 
This was a conclusion of the influential M3 forecasting competition held in 1999 \citep{Makridakis2000-ih}. So, complex methods are often viewed poorly in this field, and this has been especially true for neural networks (NN) and other Machine Learning (ML) techniques. 
In particular, AutomatANN, the only NN variant that participated in the M3, could not outperform statistical approaches that mostly headlined the rankings. 
NNs did also not perform well in subsequent competitions, e.g., in the NN3 and NN5 forecasting competitions, which were held specifically for ML methods. In the NN3 competition \citep{Crone2011-vv}, only one participating ML method was able to outperform damped trend exponential smoothing, and none of the methods was able to outperform the Theta method, which is equivalent to simple exponential smoothing with drift \citep{Hyndman2003-kc}. Both these methods are relatively simple standard methods in time series forecasting. 

Nonetheless, over the past two decades, numerous advances have been developed on the way to uncover the true potential of NNs for time series forecasting.
Recent developments have been mainly around preprocessing techniques such as deseasonalization and detrending to supplement the NN's learning process, and novel NN architectures such as recurrent neural networks (RNN), echo state networks (ESN), generalized regression neural networks (GRNN) and ensemble architectures to uplift the constraints of the conventional NN architecture \citep{Nelson1999-bd,Zhang2005-pk,Ilies2007-ej,Rahman2016-os,Yan2012-dh, Zimmermann2012-cp}. 
Also, careful selection of network parameters with the right choice of model architectures have proven that now NNs can be a strong alternative to traditional statistical forecasting methods \citep{Adya1998-ej,Zhang1998-tq,Crone2011-vv}.

Though some works have shown that NNs can be competitive even in situations where few data is available \citep{Kourentzes2014-kl,Trapero2012-di}, they typically realize their full potential in situations with more data. 
From short individual series the amount of information that can be extracted is limited~\citep{Zhang1998-tq, Yan2012-dh}.
In such a situation, simpler, more rigid models not sensitive to noise and with reasonable prior assumptions about the data will typically perform well. Complex models, in contrast, may not have enough data to fit their parameters reliably, and without proper regularization they are in danger of overfitting, i.e., they may fit to the random noise in the training data.
On the other hand, when more data is available, more model parameters can be estimated reliably, prior assumptions about the data get less important, and more generic, complex models can be estimated without being prone to overfitting. With more data available, also the distinction between signal and noise becomes clearer and simple models will not be able to fit complex signals and they underfit. 
In non-parametric methods such as NNs, the amount of model parameters and therewith model complexity can be increased when the amount of available data increases. So, when their model complexity is controlled adequately they are suitable for both situations, though in situations with less data they have strong competition from simpler, more specialized models that make reasonable prior assumptions about the data, and modelling of prior knowledge and model complexity needs to be done carefully.
%
%
A specialty of time series forecasting is that these considerations often even hold if large amounts of data are available in a time series. The distant past is typically less useful for forecasting, as underlying patterns and relationships will have changed in the meantime, so that the amount of data with relevant characteristics for forecasting is still limited.

So, a common notion is that unless the underlying time series is very long and from a very stable system, NNs will not be able to substantially outperform simpler models, as they will not have enough data to fit complex models or they will not handle non-stationarity in the data adequately (Hyndman, 2016).

These general considerations from a univariate time series context also do not readily change with the advent of ``Big Data'', where ever increasing quantities of data are collected nowadays by many companies for the routine functioning of their businesses, for example server performance measures in computer centers, sales in retail of thousands of different products, measurements for predictive maintenance, smart meter data, etc.
This is because in a time series context, availability of more data does not usually mean that the isolated series change or contain more data, e.g., that they are longer or have a higher sampling rate, as these are determined by the application and not by capturing and storage capabilities. Instead, it means that large quantities of related, similar series are available.

So, despite Big Data being a natural contributor to forecasting by providing vast quantities of data, state-of-the-art time series forecasting techniques are yet to uncover its true potential. This is mainly because traditional univariate forecasting techniques treat each time series separately, and forecast each series in isolation. Therefore, forecasting time series in these domains with traditional univariate forecasting procedures leaves great potential for producing more accurate forecasts untapped, as a separate model is built for each time series, and no information from other series is taken into account. 

Here, a competitive advantage unfolds to forecasting models that can be trained globally across all series, where traditional univariate forecasting techniques such as ETS, ARIMA, Theta, etc., are unable to exercise. To exploit the similarities between related time series, methods to build global models across sets of time series have been introduced. For example, \cite{Hartmann2015-uu} introduce a cross-sectional regression model to sets of related time series observed at the same period of time to alleviate the presence of missing values in a single time series. Also, \cite{Trapero2015-lv} use a pooled regression model by aggregating sets of related time series to produce reliable promotional forecasts in the absence of historical sales data.
However, universal function approximation properties, i.e., capacity to estimate linear and non-linear functions \citep{Cybenko1989-fw,Hornik1991-wd}, along with the large quantities of time series data available with the exposure of Big Data, have positioned NNs as ideal candidates to exploit the information dispersed across many time series.

RNNs, and in particular Long Short-term Memory (LSTM) networks have become increasingly popular to fill this gap. They are naturally suited for modelling problems that demand capturing dependency in a sequential context, and are able to preserve knowledge as they progress through the subsequent time steps in the data. As a result, RNN architectures are heavily used in domains such as Natural Language Processing \citep{Mikolov2010-rb}, machine translation \citep{Sutskever2014-vp}, and speech recognition \citep{Graves2013-cu}, and are also gaining popularity in time series research \citep{Fei2015-vh,Pawlowski2015-zs,Lipton2015-vj,Zimmermann2012-cp}. Recently, they have proven to be very competitive in the work of \cite{Smyl2016-ee} who presented an algorithm that was able to win the CIF2016 forecasting competition for monthly time series \citep{Stepnicka2016-bu}, outperforming state-of-the-art univariate algorithms such as ETS \citep{Hyndman2008-yd}, BaggedETS \citep{Bergmeir2016-zk}, Theta \citep{Hyndman2003-kc}, and ARIMA \citep{Box2015-bz}.





When building such global models for a time series database, now the problem arises that these global models are potentially trained across disparate series, which may be detrimental to the overall accuracy. 
We propose to overcome this shortcoming by building separate models for subgroups of time series. The grouping can be based on additional domain knowledge available, or, in the absence of such a natural grouping, we propose a fully automatic mechanism that works on time series databases in general, which accounts for the dissimilarities in a set of time series. The proposed methodology can be generalised to any RNN variant such as LSTMs, Gated Recurrent Units (GRUs), and others. To assess our methodology, we use LSTMs, a promising RNN variant, which is heavily used in the sequence modeling paradigm. In particular, we propose to augment the original RNN forecasting framework developed by \cite{Smyl2016-ux} with a time series clustering schema that improves the capability of the RNN base algorithm by exploiting similarities between time series.

Specifically, our proposed method initially discovers clusters of similar series from the overall set of time series, as an augmentation step to exploit the similarity between time series. We propose a feature-based clustering approach using a set of interpretable features of a time series to obtain meaningful clusters. Firstly, we extract the respective features from a time series using the method proposed by \cite{Hyndman2015-am}. Then, the ``Snob'' clustering algorithm, a mixture model based on the Minimum Message Length (MML) concept, introduced by \cite{Wallace2000-af}, is applied to the extracted feature vector, to obtain the clusters. Once we distinguish the time series based on their feature properties, for each cluster of time series, we build a separate RNN predictive model. We stabilize the variance of the series, and then we handle seasonality by a two-staged approach including deterministic deseasonalization of the series and seasonal lags. The trend is handled by a window normalization technique.
Our results show that prior subgrouping of time series is able to improve the performance of the baseline RNN model in many situations.


The rest of the paper is organized as follows. In Section~\ref{sec:lit}, we provide a brief review on the evolution of neural networks in time series forecasting and an overview of time series clustering approaches. In Section~\ref{sec:methods}, we discuss the proposed methodology in detail. Section~\ref{sec:experiments} presents the experimental setup and the results, and Section~\ref{sec:conc} concludes the paper.

\section{Related work}
\label{sec:lit}

In the following, we discuss related work in the areas of forecasting with NNs and clustering methods for time series.

\subsection{Forecasting with Neural Networks}

The powerful data-driven self-adaptability and model generalizability enable NNs to uncover complex relationships among samples and perform predictions on new observations of a population, without being constrained by assumptions regarding the underlying data generating process of a dataset. These promising characteristics are further strengthened by the universal function approximation properties that NNs possess \citep{Cybenko1989-fw,Hornik1991-wd}. Therefore, NNs are popular for classification and regression, and also in time series forecasting when external regressors and additional knowledge is available.
In pure univariate time series forecasting, over the past two decades, NN architectures have been advocated as a strong alternative to traditional statistical forecasting methods \citep{Zhang1998-tq}.
Researchers have been increasingly drawing their interest towards developing and applying different NN models for time series forecasting. This includes multi-layer perceptrons (MLP), GRNNs, ensemble architectures, RNNs, ESNs and LSTMs, while MLPs are being the most widely used NN variant for time series forecasting thus far. For a detailed description of the MLP architecture and its widespread applications employed in time series forecasting see \cite{Zhang1998-tq}.

\cite{Yan2012-dh} highlights several design implications of the MLP architecture for time series forecasting, such as a large number of design parameters, long training time, a potential of the fitting procedure to suffer from local minima, etc. To overcome these shortcomings, those authors introduce GRNN, a special type of neural network that contains a single design parameter and carries out a relatively fast training procedure compared to vanilla MLP. Also, they incorporate several design strategies (e.g., fusing multiple GRNNs) to automate the proposed modelling scheme to make it more desirable for large-scale business time series forecasting. 

There has been an increasing popularity of NNs with ensemble architectures for time series forecasting \citep{Crone2011-vv,Rahman2016-os,Barrow2016-yv,Kourentzes2014-bm,Ben_Taieb2011-iu,Zhang2001-qz,Barrow2010-fe}. In general, ensembles generate multiple versions of predictors, which when combined provide more stable prediction models with less variance and higher generalizability. Many variations of ensemble architectures have been employed in time series forecasting, such as model stacking \citep{Rahman2016-os,Crone2011-vv}, bagging and boosting \citep{Barrow2016-yv, Kourentzes2014-bm}. They address various aspects of data, parameter, and model uncertainties of the NN models.

More recently, RNN architectures are increasingly gaining interest in the time series forecasting community \citep{Zimmermann2012-cp, Fei2015-vh, Lipton2015-vj}, as they have properties that make them suitable for forecasting. We provide a detailed overview of RNN algorithms in Section~\ref{sec:rnn} and discuss their superior suitability over conventional MLP architectures to model sequential data such as time series. For example, an ESN \citep{Ilies2007-ej}, a special variant of RNN, was able to perform best among AI contenders in the NN3 competition \citep{Crone2011-vv}, and LSTMs, another popular form of RNNs that are specifically introduced to alleviate the limitations of vanilla RNNs, have been successfully applied in several time series forecasting applications \citep{Tian2015-nz,Duan2016-ta, Lee2015-pu}. 

However, despite the architectural suitability and increasing developments in the use of RNN architectures for time series forecasting, many existing studies exhibit design weaknesses, so that the forecasting community remains hesitant. This includes lack of empirical evidence and absence of evaluation metrics and standard benchmarks that are widely accepted in the forecasting community \citep{Armstrong2006-zj, Scott_Armstrong2001-as}.

\subsection{Time Series Clustering}
\label{sec:tsc}


We distinguish three main approaches to time series clustering \citep{Warren_Liao2005-vx}, namely algorithms that work directly with distances on raw data points (distance-based), indirectly with features extracted from the raw data (feature-based), or indirectly with models built from the raw data (model-based).

The performance of distance-based clustering approaches depends greatly on the particular distance metric used.
Defining an adequate distance measure for raw time series can be a challenging task, and has to consider noise, different length of the series, different dynamics, different scales, etc. Also, many such measures mostly focus on the shape of the respective time series \citep{Aghabozorgi2015-sk}, and they may yield unintuitive results and limited interpretability \citep{Wang2006-ak}.

Therefore, we focus on feature-based clustering techniques, which, instead of capturing similarity of point values using a distance metric, use sets of global features obtained from a time series to summarize and describe the salient information of the time series. Feature-based approaches can be more interpretable and more resilient to missing and noisy data \citep{Wang2006-ak}. The feature-based clustering is comprised of two stages, namely a feature extraction phase and the clustering phase, for which standard clustering approaches can be used.

In terms of feature extraction, there is a lot of work present in the literature investigating the use of features of a time series as a data-mining tool for extracting useful patterns \citep{Nanopoulos2001-gz, Wang2006-ak, Fulcher2014-de,Morchen2003-ew}. 
The two main approaches that we identify are to either extract as many features as possible or to use a limited set of carefully selected features that are interpretable and have a justification in the application case.
%
%
Following the first approach, \cite{Fulcher2014-de} introduce an automated feature construction process, using a large database of time series operations, which allows those authors to construct over 9000 different features, representing a wide range of properties of time series. 
As such a large amount of features seems not practical for our purpose, and limiting the amount of features is desirable, in our proposed framework we follow the second approach, and use a set of self-describable features proposed by \cite{Hyndman2015-am} to obtain a meaningful division of clusters. 
These suggested features are designed to capture the majority of the dynamics that can be observed in time series common in many application cases, such as trends, seasonality, autocorrelation, etc. 
Table \ref{tab:anomal} summarizes the respective feature vector that is extracted from an individual time series. In our work, we use the implementation available in R, in the \verb|tsmeasures| function from the \verb|anomalous-acm| package~\citep{Hyndman2015-am}. 

The feature extraction phase is then followed by a clustering phase that discovers the optimal grouping between the time series by applying a conventional clustering algorithm to the extracted feature vector. Again, a host of different clustering methods exist, an overview gives, e.g., \cite{Berkhin2006-mf}. 
We use a mixture-model-based clustering algorithm called ``Snob'', which we discuss in detail in Section~\ref{sec:snob}. The algorithm is not as popular as other clustering techniques, but it has a couple of advantages, e.g., it is capable of determining the optimal number of clusters in a dataset, which is an essential feature on the way towards a fully autonomous forecasting framework.

\begin{table*}
\centering
\begin{tabular}{rl}
	\hline
	            Feature    &Description \\ \hline
	             Mean     &Mean\\
	          	 Var &     Variance\\
	             ACF1 &    First order of autocorrelation \\
	      		Trend &    Strength of trend  \\
	          Linearity &  Strength of linearity \\
	      	 Curvature &   Strength of curvature \\
	    	 Season &    Strength of seasonality  \\
	      		 Peak  &   Strength of peaks   \\
	            Trough &   Strength of trough  \\
	     	 Entropy   &   Spectral entropy  \\
	        Lumpiness &   Changing variance in remainder   \\
	          Spikiness &   Strength of spikiness  \\
				Lshift &   Level shift using rolling window  \\
	  			Vchange &   Variance change   \\
	             Fspots &   Flat spots using disretization  \\
	             Cpoints &  The number of crossing points  \\
	            KLscore &   Kullback-Leibler score  \\
	           Change.idx &   Index of the maximum KL score  \\\hline
\end{tabular}
\caption{Summary of features extracted from a time series, following \cite{Hyndman2015-am}} 
\label{tab:anomal}
\end{table*}

\section{Methods}
\label{sec:methods}

In this section, we describe in detail the different parts of our proposed methodology. We first provide a brief introduction to our base algorithm, i.e., to RNNs and LSTMs. Afterwards, we explain the time series clustering method that is utilized to group sets of similar time series in the absence of other groupings. 
Finally, we discuss the time series preprocessing techniques used in our forecasting framework. This includes variance stabilization, deseasonalization, and a local normalization in a sliding window approach that structures the training data.


\subsection{Recurrent Neural Networks}
\label{sec:rnn}

RNNs are a special type of NNs that are suited to model sequences of variable lengths \citep{Elman1990-my}. In addition to the standard input and output layer, RNNs contain special hidden layers that are composed of recurrently connected nodes. These hidden layers are commonly referred to as memory states, as they enable the networks to preserve the sequential information and persist the knowledge acquired from subsequent time steps. The past information is retained through a feedback loop topology, where as a part of the input of the current step, the RNN uses the output of the previous time step during the network training. In effect, this recurrent model enables the network to take the previous values into account. There is a whole family of RNN design patterns available to model sequences, which can be distinguished by the type of recurrent architecture they use, i.e., recurrent connections between hidden units, or recurrent connections only from the output to the hidden units. For more information on RNN design patters, we refer to \cite{Goodfellow2016-yn}.



  
\begin{figure}[htb]
  \begin{center}
    \includegraphics[width=0.8\textwidth]{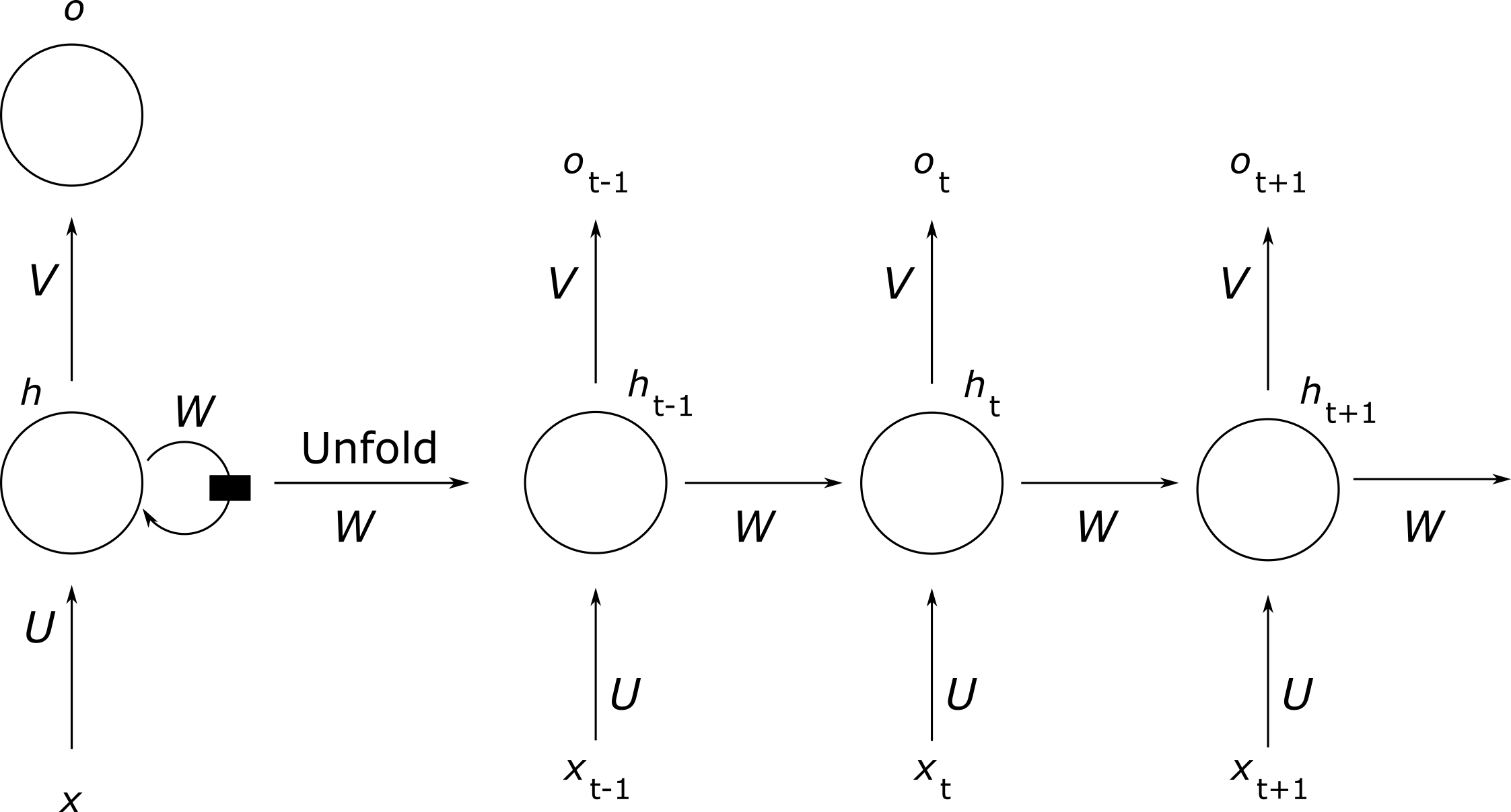}
    \caption{An unrolled recurrent neural network in time, with the shared weights of $U,$ $V,$ and $W.$}
  \label{fig:rnn}
  \end{center}
\end{figure}

Figure~\ref{fig:rnn} shows an example of an RNN unfolded in time, which unrolls the feedback loop to expand the complete sequence of the NN in time. The matrices $U$ $\in$ ${\Bbb R^{m \times n}}$, $W$ $\in$ ${\Bbb R^{n \times n}}$, and $V$ $\in$ ${\Bbb R^{n \times k}}$ define the shared weights of the RNN, where the inputs to the hidden connections are parameterized by the weight matrix $U,$ hidden-to-hidden recurrent connections are parameterized by the weight matrix $W,$ and hidden-to-output connections are parameterized by the weight matrix $V.$ Moreover, $n$, $m$, and $k$ represent the sizes of state, input, and output vectors, respectively.
In this model, $x_t$ $\in$ ${\Bbb R^{m}}$ denotes the input at time step $t$, $h_t$ $\in$ ${\Bbb R^{n}}$ denotes the hidden state at time step $t$, $h_{t-1}$ $\in$ ${\Bbb R^{n}}$ denotes the prior state at time step (${t-1}$), and $o_t$ $\in$ ${\Bbb R^{k}}$ denotes the output at time step $t$. Note that $h_t$ represents the ``memory'' of the network at time step $t$, which is computed based on the current input $x_t$ and the previous hidden state $h_{t-1}$ at time step (${t-1}$). In other words, this is the overall knowledge and reasoning accumulated by the network based on the previous data. The hidden state $h_t$ and output $o_t$ at time step $t$ can be formally defined as follows:
\begin{align*}
  h_t             & = f_{\theta}(Ux_t + Wh_{t-1})\\
  o_t             & = softmax(Vh_t)
\end{align*}
Generally, $f_{\theta}$ is a non-linear function such as, e.g., \textit{tanh}, or \textit{rectified linear units} \citep[ReLu;][]{Le2015-vb}. Whereas, \textit{softmax} is heavily used in the Natural Language Processing (NLP) context to predict a categorical output $o_t$ for a given sequence at time step $t$. RNNs often use backpropagation through time \citep[BPTT;][]{Williams1995-vl} and real time recurrent learning \citep[RTRL;][]{Robinson1987-lk} as the learning algorithms. These are extensions of the backpropagation algorithm, which unrolls the network through time to propagate the error sequentially. 

RNNs also have a close connection with classical time series forecasting methods, and can be seen as an extension of statistical ARMA models. For example, consider a general class of an ARMA($p$,$q$) model, defined as:

\begin{align*}
     x_{t} =  c+ \sum_{i=1}^{p}\phi_{i}x_{t-i} + \sum_{j=1}^{q}\theta_{j}\varepsilon_{t-j} + \varepsilon_{t}   
\end{align*}

Here, $p$ denotes the order of lags in the autoregressive model and $q$ denotes the order of error terms in the moving average model. Moreover, $c$ is a constant and $\varepsilon_{t}$ is a white noise error term. Likewise, with time series data, NNs can construct non-linear autoregressive models using time lagged observations and explanatory variables. Also, the recurrent nature of an RNN enables the model to construct weighted moving average terms of past forecasting errors using the information acquired from (${t-1}$) time steps and actual values at time step (${t}$), e.g., through BPTT. Therefore, a non-linear generalization of the linear ARMA model can be expressed as follows:

\begin{align*}
     y_{t} =   g(y_{t-1},y_{t-2},\cdots, y_{t-p},\varepsilon_{t-1}, \varepsilon_{t-2}, \cdots ,\varepsilon_{t-q}) + \varepsilon_{t}
\end{align*}

Here, $g$ represents an unknown smooth function and $\varepsilon_{t}$ is a white noise error term. For a detailed derivation and non-linear approximation of feedforward networks and fully interconnected RNNs to $g$, we refer to \cite{Connor1992-wm, Connor1994-vs}. Hence, an unrolled RNN in time approximates the statistical ARMA framework, in the sense that it is a non-linear approximation of autoregressive moving average models, which can be expressed as a NARMA($p$,$q$) model.
We note that due to their internal state and their ability to capture long-term dependencies, one could also argue that RNNs are related to state-space models and generalize exponential smoothing. However, to the best of our knowledge this analogy has not been discussed extensively in the literature.


Several properties distinguish RNNs from traditional MLPs, which makes them more appropriate for sequence modelling. As discussed earlier, the output feedback topology held by RNNs provides a notion of memory to the system, while accounting for non-trivial dependencies between the records. This task is handled by the internal state vector ($h_t$) that acts as a memory to persist the past information, while accounting for the dependencies that are longer than a given training input window. Whereas conventional feed-forward models such as MLPs are unable to bring such notion of memory to their network training process. This is essential while modelling sequences, and in particular time series, where the order of the observations is important. 
Also, RNN architectures support modelling of variable length input and output vectors, which makes them more suitable to sequence modelling compared to traditional MLPs, which are often constrained by a fixed length of inputs and outputs. Furthermore, unlike in vanilla MLPs, which use different weight parameters at each hidden layer, RNNs share the same set of weight parameters, i.e., $U,$ $V,$ and $W,$ during the network training process. This reduces the total number of parameters to be learned by the algorithm and decreases the risk of overfitting.

Even though RNN architectures are quite capable of capturing short-term dependencies in sequential data, they often have difficulties in learning long-term dependencies from distant past information. This is caused by the \textit{vanishing gradient problem} \citep{Hochreiter1991-en,Bengio1994-oq}, which is a well-known constraint in gradient based learning algorithms. Generally, gradient-based learning techniques determine the influence of a given input, based on the sensitivity of network parameters on the output. For example, as the length of a sequence grows, the corresponding error gradient is propagated through the network many steps. As a result, the gradient decays exponentially as it progresses through the chain, leaving a small impact on the output from the initial elements of the sequence. This decreases the information retention ability of RNNs, while failing to capture the potential impact from initial inputs to the network output.

\subsection{Long Short-Term Memory Networks}
\label{sec:lstm}

LSTM networks were introduced by \cite{Hochreiter1997-wo}, to address the long term memory shortage of vanilla RNNs. The LSTM extends the RNN architecture with a standalone memory cell and a gating mechanism that regulates the information flow across the network. The gating mechanism is equipped with three units, namely: input, forget, and output gate. This mechanism cohesively determines which information to be persisted, how long it is to be persisted, and when it is to be read from the memory cell.

\begin{figure}[htb]
  \begin{center}
    \includegraphics[width=0.6\textwidth]{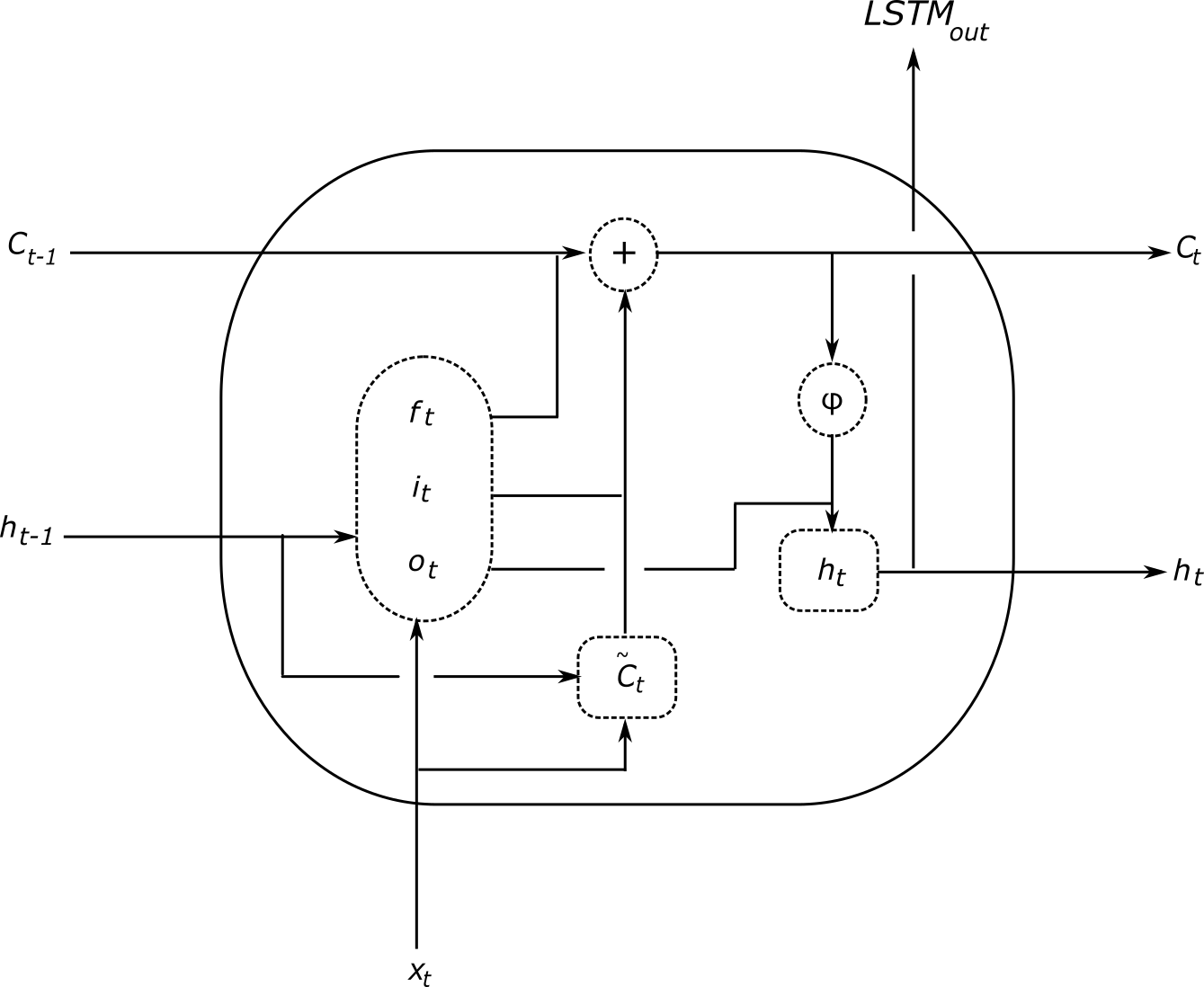}
    \caption{Basic architecture of an LSTM memory block with three gated layers: forget gate $\protect{f_t}$, input gate $\protect{i_t}$, and output gate $\protect{o_t}$, controlling the activation of cells $\protect{C_{t-1}}$ and $\protect{C_t}$}.
  \label{fig:lstm}
  \end{center}
\end{figure}

As a result, LSTMs are capable of retaining key information of input signals, and ignore less important parts. This memory cell has a recurrently self-connected linear unit called ``Constant Error Carousel'' (CEC), which contains a state vector ($C_{t}$) that allows to preserve dependencies for the long-term. Consequently, in contrast to vanilla RNNs, LSTMs preserve information and propagate errors for a much longer chain in the network, and overcome the vanishing gradient problem. In fact, LSTMs possess the ability of remembering over 100 steps of a sequence \citep{Langkvist2014-rn}. Figure~\ref{fig:lstm} illustrates the basic structure of an LSTM memory block with a one cell architecture \citep[following][]{R2RT_Blog2016-ya}. 
In the figure, $x_t$ $\in$ ${\Bbb R^{m}}$ denotes the input at time step $t$, $C_{t-1}$ $\in$ ${\Bbb R^{n}}$ and $C_{t}$ $\in$ ${\Bbb R^{n}}$ denote the cell state at time steps $t$ and (${t-1}$), while $h_{t}$ $\in$ ${\Bbb R^{n}}$ and $h_{t-1}$ $\in$ ${\Bbb R^{n}}$ correspond to the output at time steps $t$ and (${t-1}$), respectively. Here, $m$ represents the size of input vector, while $n$ denotes the size of the memory cell. In general, the short-term memory of LSTMs is provided by the $h_{t}$ state, while the $C_{t}$ state enables to retain long-term dependencies.


Also, to distinguish the self-contained memory cell CEC of the LSTM from the conventional state ($h_{t}$), we refer to it as $C_{t}$. The forget gate $f_{t}$ takes $x_t$ and $h_{t-1}$ as inputs to determine which information to be retained in $C_{t-1}$. The gate activation functions $i_t$, $o_t$, and $f_t$ are usually sigmoid layers, so that the output is projected to a value between zero and one for each value in $C_{t-1}$, describing the scale of information retention, i.e.,  a ``zero'' output represents the complete expunge of a value from the memory cell, while a ``one'' represents the complete retention of that value in the memory cell. Meanwhile, the input gate $i_t$ is accompanied with a sigmoid layer that uses $x_t$ and $h_{t-1}$ to ascertain the values to be summed by addition to $C_{t}$. Additionally, a non-linear layer ${\phi}$ (e.g., \textit{tanh}) is also introduced to generate a vector of candidate values, denoted as ${\tilde{C_{t}}}$, to update the state of $C_{t}$. The output gate $o_{t}$ regulates the output values of an LSTM cell, based on the updated state of $C_{t}$. Likewise, as in the forget and input gates, the output gate is a sigmoid layer to filter the output. Correspondingly, the updated cell state $C_{t}$ is fed into a \textit{tanh} layer (${\phi}$), which scales down the vector to a value between (-1) and (+1). This is then multiplied element-wise by the output of the sigmoid layer to compute the final cell output $h_{t}$ at time step $t$. The aforementioned process can be formally defined by the following recursive equations:

\begin{align*}
  i_t             & = {\sigma}(W_{i}{\cdot}h_{t-1} + U_{i}{\cdot}x_{t} + b_{i})\\
  o_t             & = {\sigma}(W_{o}{\cdot}h_{t-1} + U_{o}{\cdot}x_{t} + b_{o})\\
  f_t             & = {\sigma}(W_{f}{\cdot}h_{t-1} + U_{f}{\cdot}x_{t} + b_{f})\\
  \tilde{C_{t}}	  & = {\phi}(W_{c}{\cdot}h_{t-1} + U_{c}{\cdot}x_{t} + b_{c})\\
  C_t             & = f_{t}\, {\odot} \,C_{t-1} \,  + i_{t}\, {\odot}\, \tilde{C_{t}}\\
  h_t 			  & =  o_t \,{\odot}\,{\phi}(C_t) \\
  LSTM_{out}      & = h_t  
\end{align*}

%
Here, ($W_{i}$, $W_{o}$, $W_{f}$, $W_{c}$) $\in$ ${\Bbb R^{n \times n}}$ represent the weight matrices of forget gate, input gate, memory cell state, and output gates respectively. Also, ($U_{i}$, $U_{o}$, $U_{f}$,$U_{c}$) $\in$ ${\Bbb R^{m \times n}}$ denote the corresponding input weight matrices. Biases of the respective gates are ($b_{i}$, $b_{o}$, $b_{f}$, $b_{c}$) $\in$ ${\Bbb R^{n}}$, while $\odot$ denotes the element-wise multiplication operation. In these equations, $\sigma$ represents the standard logistic sigmoid activation function, and $\phi$ stands for the hyperbolic tangent function, i.e., \textit{tanh}. These are defined by the following equations respectively:

\begin{align*}
	 {\sigma}(x) =  \frac{\mathrm{1}}{\mathrm{1} + e^{-x}}\\
	 {\phi}(x) =  \frac{e^{x} - e^{-x}}{e^{x} + e^{-x}}
\end{align*}



Several variants to the originally proposed algorithm can be found in the literature. E.g., LSTM with ``peephole connections,'' introduced by \cite{Gers2000-hc}, is one of the popular variants that allows LSTM gates to examine the state of their memory cell, before updating their states. Let the peephole weight matrices of input, output and forget gates be defined as ($P_{i}$, $P_{f}$, $P_{o}$) $\in$ ${\Bbb R^{n \times n}}$. Then, the equations of the basic LSTM can be rewritten as follows:

\begin{align*}
   i_t             & = {\sigma}(W_{i}{\cdot}h_{t-1} + U_{i}{\cdot}x_{t} + P_{i}{\cdot}C_{t-1} + b_{i})\\
   f_t             & = {\sigma}(W_{f}{\cdot}h_{t-1} + U_{f}{\cdot}x_{t} + P_{f}{\cdot}C_{t-1}+ b_{f})\\
   \tilde{C_{t}}   & = {\phi}(W_{c}{\cdot}h_{t-1} + U_{c}{\cdot}x_{t} + b_{c})\\
   C_t             & = f_{t}\, {\odot} \,C_{t-1} \ + \, i_{t}\, {\odot}\, \tilde{C_{t}}\\
   o_t             & = {\sigma}(W_{o}{\cdot}h_{t-1} + U_{o}{\cdot}x_{t} + P_{o}{\cdot}C_{t}+ b_{o})\\
   h_t 			   & = o_t \,{\odot}\,{\phi}(C_t)\\      
   LSTM^{peephole}_{out}      & = h_t
\end{align*}

Since LSTM is an RNN variant, the sequence of the original time series is relevant and needs to be preserved during training. All training patches relevant to a particular time series are read as one sequence. 
Therefore, the LSTM state needs to be initialized for each series. Typically a vector of zeros is used, but there are other possibilities.
As a result of this state transition, the trained LSTM network with a set of fixed weight vectors can still show different predictive behaviour for different time series, i.e., in a trained network, a particular time series is represented by the composition of weight vector and its corresponding internal state \citep{Prokhorov2002-sh,Smyl2016-ux}. 

In this study, we use the Microsoft Cognitive Toolkit (CNTK), an open-source NN toolkit \citep{Seide2016-vu}, to implement the LSTM. 
As our base learning algorithm, we use an LSTM with peephole connections, which is followed by an affine neural layer (excluding the bias component) to project the LSTM cell output to the dimension of the intended forecast horizon, i.e., the dimension of this fully connected neural layer equals the size of the output window. 
%
We use L2-norm as our primary loss function to train the LSTM, which essentially minimizes the sum of squared differences between the target values and the estimated values.

\subsection{The ``Snob'' clustering method}
\label{sec:snob}

``Snob'' is a mixture model algorithm, which is based on the MML concept, a Bayesian point estimation technique that accounts for the highest posterior probability distribution of each cluster \citep{Wallace1994-lu,Wallace2000-af}. For example, after applying the Bayes' theorem, the posterior probability of a hypothesis/theory $H$, given the data $D$ can be formally written as follows: 

\begin{align*}
	  Pr(H \& D) = Pr(D\mid H){\cdot}Pr(H)
\end{align*}

Here, $Pr(H \& D)$ refers to the joint probability of $H$ and $D$. According to Shannon's information theory \citep{Shannon1948-nk}, the optimal length of the message, which describes an event $E$ is given by $-log(Pr(E))$. The above joint probability can be rearranged as a two part message length as follows, after applying Shannon's information theory:


\begin{align*}
	 I(H \& D)= \underbrace{I(H)}_{1} + \underbrace{I(D\mid H)}_{2} 
\end{align*}

Here, part(1) represents the optimal length of the message, which encodes the theory ($H$) that describes the cluster structure of each class. This contains the number of clusters, parameter distribution of each cluster, and proportion of each cluster to the entire population. Whereas part(2) denotes the optimal encoding length of the message that describes the data ($D$), given the theory ($H$). This represents the cluster assignments of each data instance and its corresponding feature vector. In essence, Snob uses the objective function of MML, i.e., $I(H \& D)$ or the total message length, to choose the best model that describes the given data.

Following the MML strategy, Snob evaluates different cluster assignments by calculating the total message length, which is the objective function to be minimized in each iteration. As a result, unlike in, e.g., the traditional K-means clustering algorithm, Snob is able to determine an optimal number of clusters in the dataset autonomously, i.e., the number of clusters doesn't have to be specified in advance. In addition to this, Snob is indifferent to scaling of the attributes and can handle attributes with different distributions and combinations of categorical and numerical attributes. As we only consider numerical attributes in this work, we limit Snob to normally distributed attributes so that it is effectively a Gaussian mixture model.

\subsection{Moving Window Approach}
\label{sec:mva}



The proposed moving window approach follows the Multi-Input Multi-Output (MIMO) strategy of forecasting that models a multiple input and output mapping, while preserving the stochastic dependencies between predicted values. 
While RNNs can be operated with one input at a time, the internal state of the network then needs to memorize all relevant information. Using an input window relaxes this requirement and allows the network to also operate directly with lagged values as inputs. Furthermore, \cite{Ben_Taieb2011-iu} discuss the benefits of applying a MIMO strategy over a single-output forecasting strategy (recursive strategy) in multi-step forecasting. There, those authors highlight that accuracy of the latter approach is affected by its recursive nature, and errors are accumulated at each forecasting step.

\begin{figure}[htb]
  \begin{center}
    \includegraphics[width=0.9\textwidth]{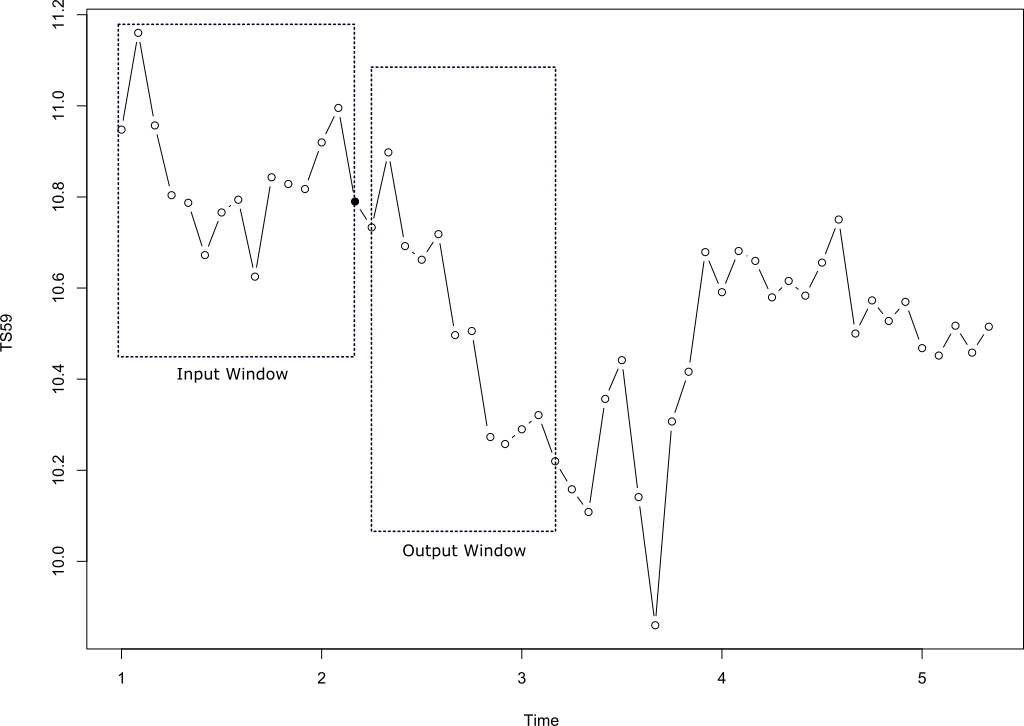}
    \caption{An example of applying the moving window approach to the series TS59 of the CIF2016 dataset. The filled dot marks the last value of the input window.}
  \label{fig:window}
  \end{center}
\end{figure}

In this work, we apply the moving window method to a fully preprocesses time series as discussed in the following sections, as follows. 
%
At first, a time series of length \textit{tsLength} is converted to patches of length (\textit{outputSize+inputSize}). In total, there are (\textit{tsLength--outputSize--inputSize}) such patches. Here, \textit{outputSize} refers to the length of the output window (i.e., the intended forecasting horizon), while \textit{inputSize} represents the length of the input window used in each frame. Figure~\ref{fig:window} illustrates the procedure with an example of applying the moving window approach to series TS59 of the CIF2016 dataset. 
The training dataset is generated by iterating the above process until the last point of the input window is positioned at (\textit{tsLength-outputSize-1}), i.e., the last output window of the series is reserved for validation and not used for training. 
For the validation, forecasts for this last output window are produced. Due to the recursive nature of the process, also for the validation we need to iterate through the whole time series (called the ``warm-up'' in this case), analogous to the training phase.

As to concrete choices of the size of input and output windows, the output window size is largely determined by the required forecast horizon. As a heuristic, we then currently choose the input window size in relation to the output window size and the seasonality. In particular, $\textit{inputSize} = 1.25\cdot max(\textit{outputSize}, seasonal\_period)$. That is, we choose the input window larger (with a factor of 1.25 in our experiments) than the output window size or the length of the seasonal period, whichever is larger. 
This choice is rather empirical, and it seems adequate for situations where a full period can be captured by relatively few data points. In situations where, e.g., a daily series with a yearly seasonality is to be forecasted, the method would likely benefit from more sophisticated feature or lag selection. 
Also, in situations where time series are very short, a smaller input window size may need to be chosen.


\subsection{Modelling seasonality}
\label{sec:tsd}




%
%
%
Early studies suggest that NNs are suitable to effectively model the underlying seasonality and cyclical patterns in time series due to their universal function approximation properties, i.e., their capacity to estimate linear and non-linear functions \citep{Zaiyong_Tang1991-wl,Marseguerra1992-fp}.
However, more recently several studies argue that deseasonalizing data prior to modelling is necessary to produce accurate forecasts. In particular, \cite{Nelson1999-bd} compare the forecasts generated from NNs trained with deseasonalized data and non-deseasonalized data, using 68 monthly time series from the M-competition \citep{Makridakis1982-oz}. The results indicate that the NN trained with prior deseasonalization achieves better forecasting accuracy, in contrast to NNs trained with non-deseasonalized data. Similarly, using the NN5 competition data, \cite{Ben_Taieb2011-iu} empirically show that the resulting forecasts benefit from prior deseasonalization of the data. \cite{Zhang2005-pk} demonstrate that NNs are not capable of effectively modelling trend or seasonality directly, and emphasize that the forecasting errors can be reduced by detrending or deseasonalization of the raw time series.
%
%
%
These findings are in line with our discussions in Section~\ref{sec:intro} of data availability, model complexity and prior assumptions. Though NNs may be able to model a signal in principle, this is no guarantee that in practice they will do it accurately. 

As our method is intended to run also especially in situations where the overall amount of data is limited, we use deseasonalization techniques to remove deterministic seasonality, which may be unnecessarily difficult to learn for the RNN. Then, inclusion of seasonal lags in our rolling window procedure allows the RNN to capture the remaining stochastic seasonality. This approach is inspired by the well-known ``boosting'' ensemble technique \citep{Schapire2003-xb}, where STL deseasonalization can be seen as a weak base learner that is subsequently supplemented by the RNN.
We use seasonal and trend decomposition using loess (STL), as proposed by \cite{Cleveland1990-rc}, which is considered a robust method to decompose a time series into trend, seasonal, and remainder components. It consists of a sequence of applications of a loess smoother, making the decomposition computationally efficient, even for longer time series \citep{Cleveland1990-rc,Hyndman2014-rj}.
%
%
%
We use STL to extract a deterministic seasonality from the series after variance stabilization (see Section~\ref{sec:logtrans}), and pass on the sum of trend and reminder to the next step of data preprocessing. 
In R, STL is implemented in the \verb|stl| function from the \verb|forecast| package~\citep{Hyndman2015-vm,Khandakar2008-hd}. To obtain deterministic seasonality, we set the \verb|s.window| parameter to ``periodic''. This parameter controls the smoothness of the change of seasonal components, and setting it to ``periodic'' enforces that no change in the components is possible, so that the result is a deterministic seasonality, where all seasonal periods are identical.


We currently apply the deseasonalization procedure to all time series, regardless of whether seasonality is present in the data or not. The only rather basic test we perform for seasonality is that if series contain less than two full seasonal periods, the STL procedure will not be applicable, and then we will assume the series is non-seasonal and consequently we will not deseasonalize the series. 
In all other cases, we extract a deterministic seasonality. The assumption we currently have is that even if the deseasonalization procedure extracts a seasonal component that is not actually present in the data, this component will be small and can easily be compensated for by the second phase of the seasonality modelling, i.e., the seasonal lags of the RNN. In any way, modelling such a small artificial seasonal component will be easier than modelling of  all the deterministic seasonality in the dataset by the RNN, without any preprocessing.

Other possibilities to address this issue would be to determine seasonality with, e.g., a statistical test, and apply deseasonalization only to series where seasonality is detected. 
However, such tests are usually either based on the Autocorrelation Function (ACF) or they are model-based \citep{Hans_Franses1992-ow}, and therewith they have their own assumptions and shortcomings, and could cause problems in our non-linear, non-parametric setup.
Furthermore, following the argumentation of \cite{Hyndman2014-na}, we are not interested in uncovering an underlying data generating process, but we are concerned in improving forecasting accuracy.

%
%


Though we restrain from modelling seasonality from only part of the series, depending on the data it is possible to imagine situations where modelling the entire seasonality by the RNN will be superior to the proposed process. Performance of the current process depends on the capability of the deterministic deseasonalization procedure to extract seasonality from a single series. Accordingly, if the dataset is such that seasonality is too noisy in the single series, or the series are too short, the proposed procedure will not be applicable, and it may be beneficial to then omit the seasonality preprocessing alltogether. 
Also, the deseasonalization is less likely needed when we face a number of time series with calendar features (e.g., day of week) and/or homogeneous seasonality patterns. If time stamps are known and the series are related, e.g. describe electric load of subsections of a regions grid, the RNN will be more likely to learn the seasonality. On the other hand, if there are different and/or unknown seasonalities across the series, it is better to deseasonalize first.
In line with the discussions above, best practice would then be to fit both a model with and without deseasonalization, and then choose the better of the two models using a validation set. 



%



\subsection{Variance stabilization using power transformations}
\label{sec:logtrans}


Some works in the literature argue that, though NNs have universal approximation capabilities~\citep{Hornik1991-wd}, power transformations may make the NN learning procedure difficult by altering the original non-linearities in a time series \citep{Faraway2008-fi}. However, we stabilize the variance in our data for two reasons. As the STL decomposition method that we employ for seasonality extraction is an additive method, we need to ensure that seasonality is additive. Furthermore, we model the trend in a conservative way (see Section~\ref{sec:trend}), and stabilizing the variance enables us to model greater dynamics in the trend.

Popular transformations for variance stabilization are, e.g., the Box-Cox transform \citep{Box1964-ti} or the log transform. 
The log transform is a strongly non-linear transform and is usually used with caution, as small differences in log space may result in large differences in the original space, and therewith model fitting can yield sub-optimal results.
The Box-Cox transform is a popular more conservative approach. It is usually defined as follows:

\[
w_t = \begin{cases}
  \log(y_t), & \lambda=0;\\
  (y_t^{\lambda}-1)/\lambda, & \lambda \neq 0.
\end{cases}
\]

We see that the transform resembles, depending on its parameter $\lambda$, the logarithm or the identity in its most extreme cases ($\lambda = 0$ or $\lambda = 1$, respectively). A difficulty with this transform is the choice of $\lambda$. The only procedure we are aware of to choose $\lambda$ automatically is the procedure of \cite{Guerrero1993-hv}. 
%
%
However, in preliminary experiments we found that this procedure has its shortcomings and the parameter $\lambda$ is difficult to choose in practice, and that in fact the logarithm seems to be the better choice for our proposed method. Therefore, in our experiments, we use the log transformation to transform each time series to a logarithmic scale before it is fed into the STL algorithm. Finally, in the post-processing stage, the corresponding forecasts are back-transformed into the original scale of the time series, by taking the exponent of each generated output value. 
To avoid problems for zero values, we use the logarithm in the following way:



\[
w_t = \begin{cases}
  \log(y_t), & min(y)>\epsilon;\\
  \log(y_t+1), & min(y)\leq\epsilon;\\
\end{cases}
\]

\begin{figure}[htb]
  \begin{center}
    \includegraphics[width=0.45\textwidth]{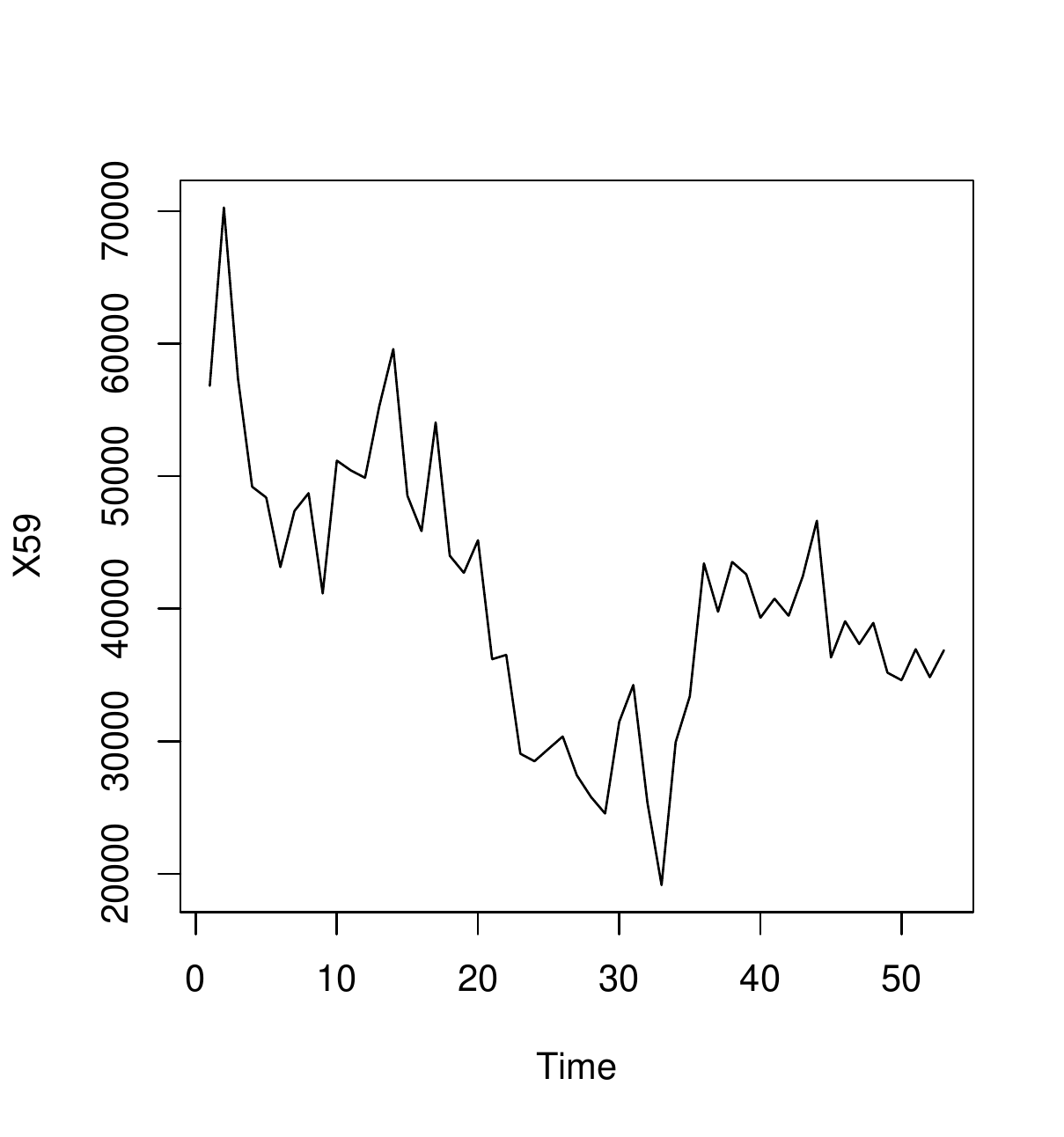}
    \includegraphics[width=0.45\textwidth]{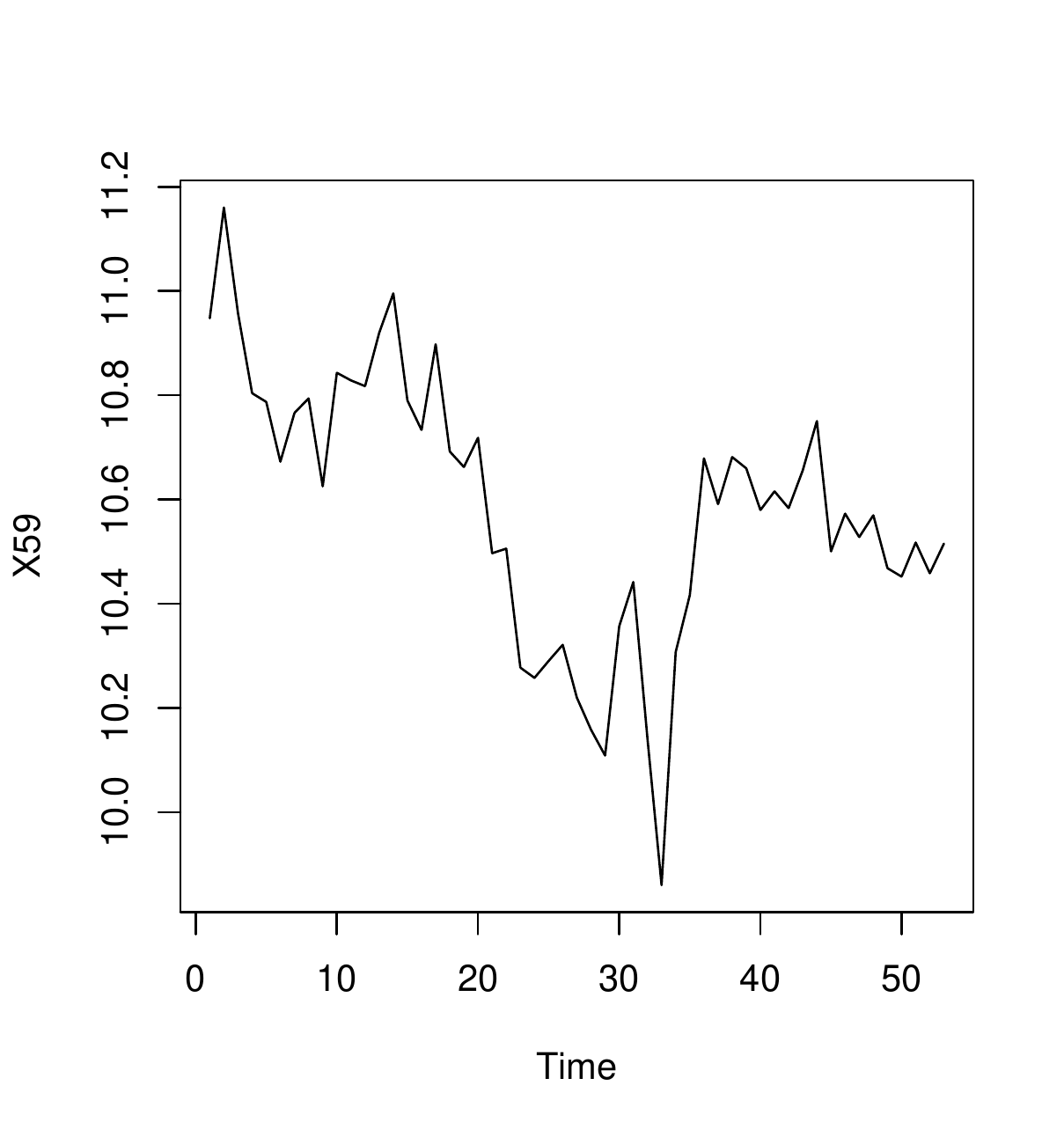}
    \caption{Series TS59 of the CIF2016 dataset, which is a monthly time series. On left is the original series, right the log transformed version.}
  \label{fig:logtrans}
  \end{center}
\end{figure}

Here, $y$ denotes a time series, and $\epsilon$ can be chosen as zero for integer time series, or a small value close to zero for real-valued time series. As an example, Figure~\ref{fig:logtrans} shows the original series and the log transformed version of Series TS59 of the CIF2016 dataset.

\subsection{Modelling trend}
\label{sec:trend}

Within the rolling window processing, the last value of the trend in each input window, provided by STL (illustrated by the filled dot in Figure~\ref{fig:window}) is used for local normalization. The trend component extracted by the STL procedure of that last value is subtracted from each data point in the corresponding input and output window. This process is applied to each input and output window combination separately. 

Modelling the trend in this way has various advantages over extracting the trend using STL and modelling it separately. In contrast to prediction of a deterministic seasonality, predicting forward the extracted trend of a time series is not trivial, so if we extracted the trend and predicted it separately, we would effectively face another non-trivial prediction problem. Instead, we use the RNN directly to predict the trend. 

As the RNN will eventually saturate, the predicted trend is limited by the bounds of the transfer function \citep{Ord2017-cj}. However, the local normalization step employed makes the network not saturate based on absolute values, but effectively it limits the steepness of the trend to the maximal steepness found in all training windows (after variance stabilization, deseasonalization). This leads to rather conservative forecasts and seems a reasonable assumption in a time series prediction scenario, where predicted exponential trends are often the source of potentially large forecasting errors. 



\begin{figure}[htb]
  \begin{center}
    \includegraphics[width=0.9\textwidth]{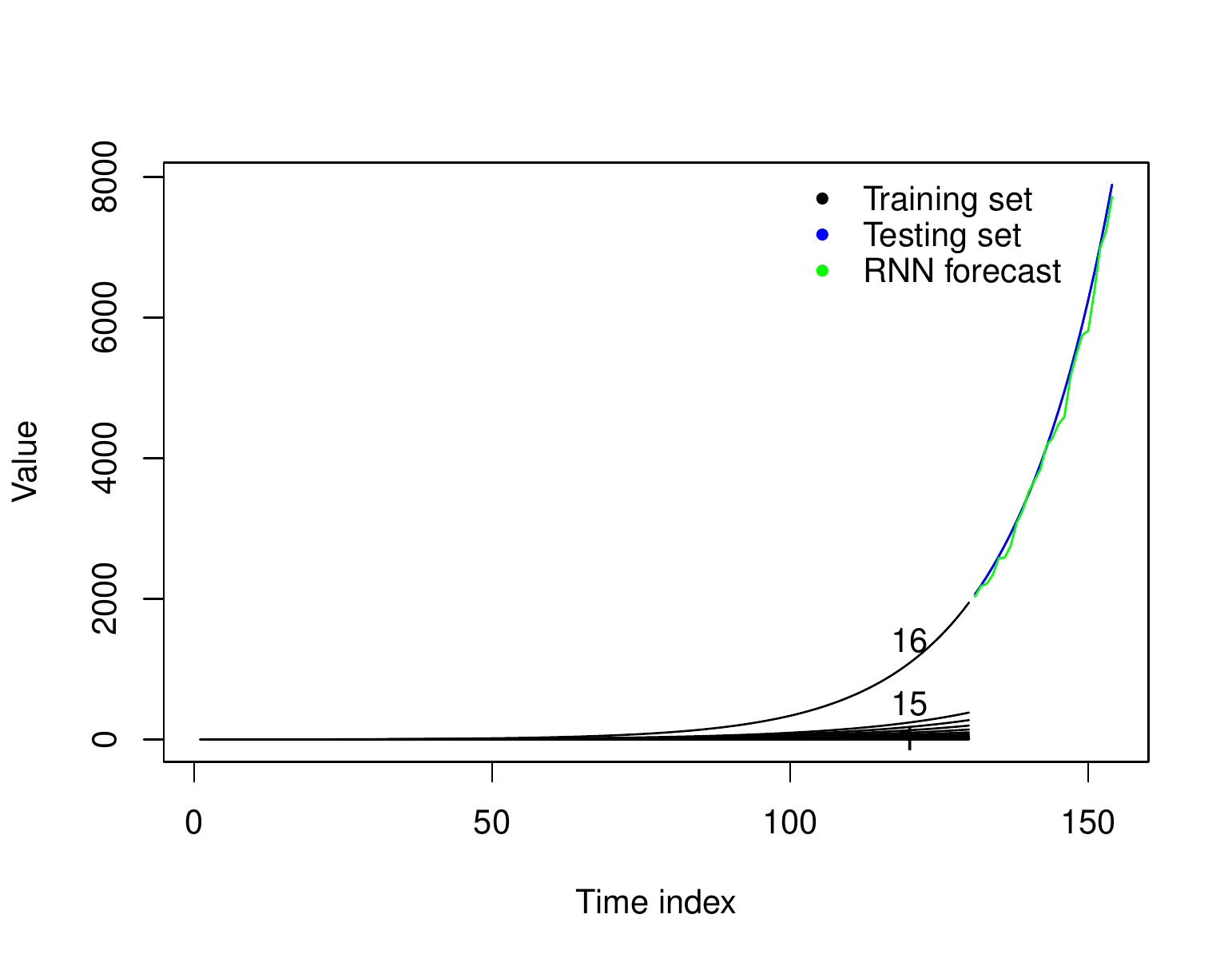}
    \caption{Set of 16 synthetically generated time series that exhibit exponential trends. Each time series contains 130 data points for training and 24 data points for testing. The steepness of the trend in each time series is elevated gradually. We see that our method is able to predict the exponential trend of the steepest time series accurately, even though the values predicted are outside the range of the training data. We note that the prediction is performed with a single output window.}
  \label{fig:trendplot}
  \end{center}
\end{figure}

To further illustrate the capability of our procedure to model trends, in Figure~\ref{fig:trendplot} we report the results of a brief experiment with simulated data. We see that in practice the procedure of variance stabilization and local normalization is able to model even fast-growing, exponential trends.

\subsection{The overall procedure}
\label{sec:overall}

To summarize, a scheme of the forecasting framework is given in Algorithm~\ref{alg:lstm}. If a partition of the time series is available in the form of additional knowledge, this partition can be considered. If no additional knowledge is present, we employ the following fully automatic procedure. As stated in Section~\ref{sec:tsc}, at first we use the ``anomalous'' feature extraction method from~\cite{Hyndman2015-am}. Then, an implementation of the Snob clustering algorithm is applied to the feature vector, to find the optimal grouping of time series \citep{Wallace2000-af}. After discovering the clusters, for each cluster of time series, the following pre-processing steps are applied to generate input data for the RNN training.
\begin{algorithm}
\begin{algorithmic}[1]
\Procedure{preprocessing}{ts, freq, input.win, output.win} 
\State ts.features $\gets$ \textbf{anomalous}(ts, freq) 
\State ts.clusters $\gets$ \textbf{rsnob}(ts.features) 
\For{i : len(ts.clusters)}
\State ts.log $\gets$ \textbf{log}(ts) 
\State [trend, seasonal, remainder] $\gets$ \textbf{stl}(ts.log, freq) 
\State ts.deseason $\gets$ [trend, remainder]
\For{i : [tsLength(ts.deseason)-output.win-1]}
\State window.frame[i] $\gets$ \textbf{rollWindow}(ts.deseason, input.win, output.win)
\State normalize.series[i] $\gets$ \textbf{normalize}(window.frame[i])
\State  [input, output] $\gets$ \textbf{get}(normalize.series[i])
\EndFor
\EndFor
\State \textbf{return} ts.series[input, output]
\EndProcedure
\caption{Generating target input files for the RNN}
\label{alg:lstm}
\end{algorithmic}
\end{algorithm}
First, we stabilize the variance, using a log transformation. Then, the series is decomposed into trend, seasonal part, and remainder using the \verb|stl| function from the \verb|forecast| package, in a deterministic setting. 
Afterwards, as stated in Section~\ref{sec:mva}, the rolling window approach, along with a local normalization technique is applied to the sum of trend and reminder, to generate the training data. 
Thereafter, for each cluster, a separate LSTM model is trained, and used for prediction.



\begin{figure}[htb]
  \begin{center}
    \includegraphics[width=0.6\textwidth]{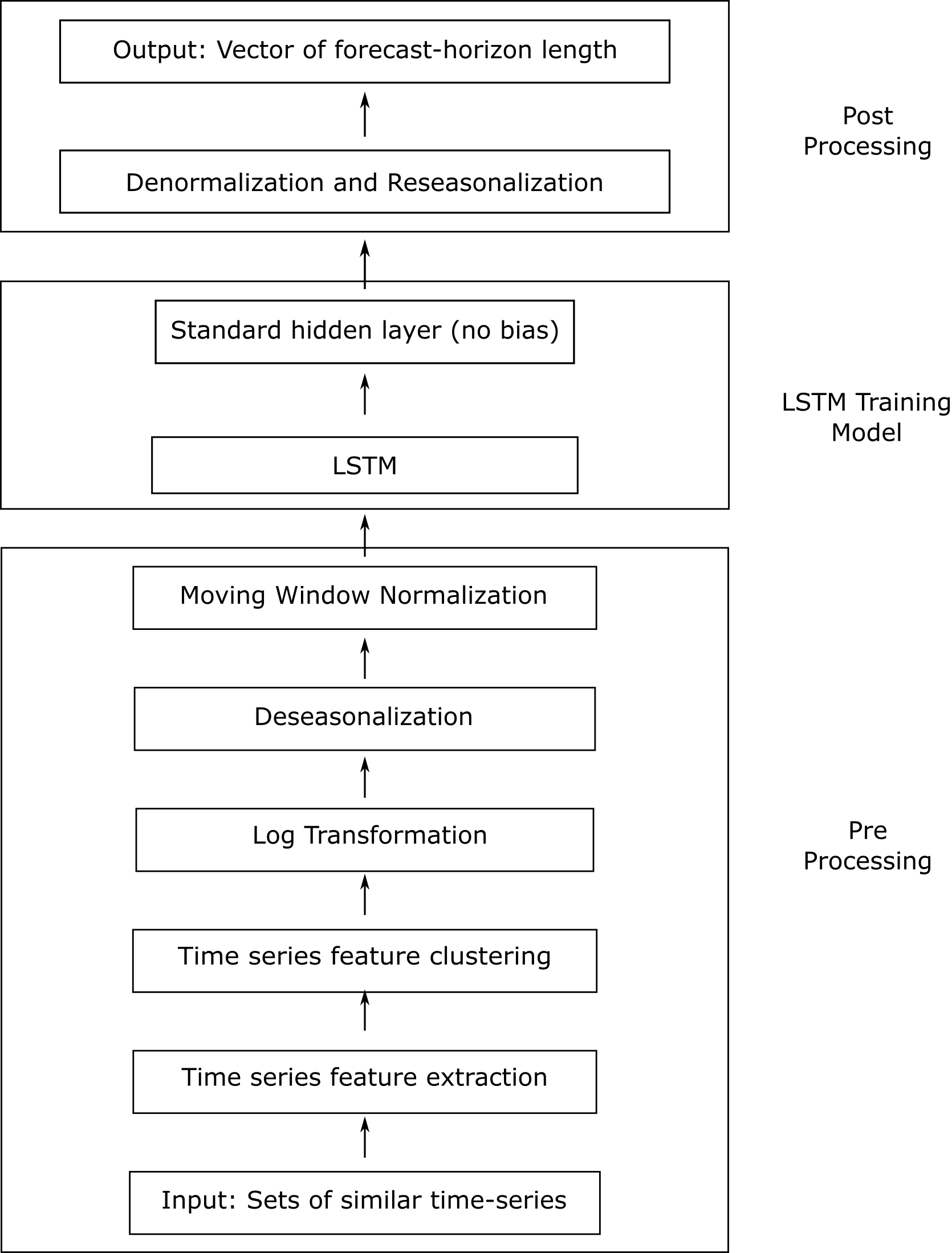}
    \caption{Proposed network architecture, which includes a pre-processing layer, LSTM training layer and a post-processing layer.}
  \label{fig:network}
  \end{center}
\end{figure}

Figure~\ref{fig:network} gives an illustration of the proposed forecasting framework. The overall model is comprised of three components, namely: 1) the pre-processing layer which consists of a clustering phase and a log-transformation, deseasonalization, and a normalization phase, 2) the LSTM training layer which consists of an LSTM layer, followed by an affine neural layer (a fully connected layer), excluding the bias component, and 3) a post-processing layer which consists of a denormalization and a reseasonalization phase to ascertain the final forecasts. The reseasonalization process includes introducing the last seasonal component to the generated forecasts. Whereas during the denormalization, the generated forecasts are back-transformed to their original scale, by adding the corresponding trend value obtained from the local normalization process, i.e, adding the last value of the trend inside an input window, and taking the exponent of the values.

\section{Experimental study}
\label{sec:experiments}

In this section, we evaluate the proposed procedure on two benchmark datasets from past forecasting competitions, namely the CIF2016 and NN5 datasets. We describe the forecasting methods and error measures used to perform the experiments, and the results obtained.

\subsection{Benchmarking datasets from forecasting competitions}

We use the publicly available datasets from the CIF2016 and NN5 forecasting competitions. These competitions were specifically organized to evaluate and compare the potential of ML techniques in handling large scale ex-ante forecasting. In fact, each dataset is comprised of similar time series, related to a certain domain. This is the main basis of using these specific public datasets, as they comply with our original hypothesis of exploiting the advantages of similar time series (unlike, e.g., the M3 competition data).

The CIF2016 competition dataset consists of monthly time series, composed of two different subgroups: series related to the banking industry and artificially generated series~\citep{Stepnicka2016-bu}. Specifically, contestants were requested to submit 12-months-ahead forecasts for 57 time series, and 6-months-ahead forecasts for 15 time series, so for a total of 72 series. The CIF2016 competition attracted participants from numerous fields of Computational Intelligence (CI), such as artificial NNs, fuzzy methods, support vector machines, decision and regression trees, etc.

The NN5 competition dataset contains 2 years of daily cash withdrawals at various automatic teller machines (ATMs) located in the UK \citep{Crone2008-ye}. In detail, 111 time series of ATMs were made available during the competition, and the participants were asked to submit the forecasts for a prediction horizon of 56 days ahead. Moreover, the NN5 competition includes various challenges of a real-world forecasting task, such as multi-step ahead forecasting, outliers, missing values, and multiple seasonalities. Similar to the CIF2016, a variety of CI solutions were presented at the competition. 

However, while primary means of ranking forecasting approaches of these competitions was among CI techniques, statistical and hybrid techniques were also permitted to submit their forecasting solutions. Therefore, we can straightforwardly evaluate our proposed approach against state-of-the-art statistical benchmarks such as ETS, ARIMA, Theta, etc. 

\subsection{Error measures}

%
To compare our proposed approach against the participants' forecasts of CIF2016 and NN5, we calculate the symmetric Mean Absolute Percentage Error (sMAPE) for every series, and then calculate an average over all the series to obtain a mean sMAPE, which is consistent with the evaluations in the competitions. To get a better understanding of outliers and single time series that may dominate this evaluation, we furthermore calculate the median and rank sMAPE over all series.

Furthermore, a host of different choices to evaluate forecasts exists in the forecasting literature, overviews and good discussions give \cite{Hyndman2006-ue} and \cite{Davydenko2013-wk}. Many popular measures such as the sMAPE have known shortcomings, such that they are skewed, lack of robustness, lack of interpretability, or are unstable with values close to zero in the original data and/or the forecasts.
%
%
To address some of these issues, we use as a second evaluation metric in our experiments the Mean Absolute Scaled Error (MASE), as proposed in \cite{Hyndman2006-ue}. The MASE is a scale independent error measure, which also offers interpretability, as it measures the forecasting accuracy relative to the average one-step na{\"i}ve forecast error, or to the na{\"i}ve seasonal forecast error, if the time series is seasonal. That is, assuming the time series are seasonal, MASE $<$ 1 means that on average the method performs better than the na\"ive seasonal forecast computed on the training data, while MASE $>$ 1 indicates that the method performs worse. We use the sMAPE and the MASE in the following definitions:


\[
\text{sMAPE} = \frac{200}{h}\sum_{t=1}^{h}\left(\frac{\left|F_t - Y_t\right|}{\left| F_t\right| + \left| Y_t\right|} \right)
\vspace{5mm}
\]
\[
\text{MASE} = \frac{1}{h}\frac{\sum_{t=1}^{h}|F_t - Y_t|}{\frac{1}{n-M}\sum_{t=M+1}^{n}|Y_{t}-Y_{t-M}|}
\vspace{5mm}
\]



Here, $Y_t$ denotes the observation at time $t$, $F_t$ is the respective forecast. Furthermore, $h$ denotes the number of data points in the test set and $n$ denotes the number of data points in the training set of a time series. The seasonal period of a time series is represented by $M$. In particular, we provide the following evaluation measures based on these primary error measures: Mean of the sMAPEs (Mean sMAPE), Median of the sMAPEs (Median sMAPE), Mean of the sMAPE ranks of each series (Rank sMAPE), Mean of the MASE (Mean MASE), Median of the MASE (Median MASE), and  Mean of the MASE ranks (Rank MASE) of each series.


%


\begin{table*}[!tb]
\centering
\begin{tabular}{lcccc}
	\toprule
	Model Parameter            			&Minimum value 			&Maximum value\\ \hline
	LSTM-cell-dimension  				&10						&80			\\
	Epoch-size   						&($n_{tr}$)					&($n_{tr}$*3)			\\
	Mini-batch-size						&2						&40			\\
	Learning-rates-per-sample			&0.001					&0.04		\\
	Maximum-epochs						&10						&40		\\
	Gaussian-noise-injection			&0.0005					&0.005\\
	L2-regularization-weight			&0.0005					&0.0008\\ \hline
\end{tabular}
\caption{Parameter value ranges used throughout the LSTM learning process, represented by the respective minimum and maximum values. Here, $n_{tr}$ denotes the number of training examples in a training file.}
\label{tab:parametergrid}
\end{table*}


\subsection{Hyperparameter selection and compared methods}


The LSTM has various hyper-parameters. We tune these with a Bayesian global optimization methodology \citep{Snoek2012-ko}, which uses Bayesian inference and a Gaussian process to autonomously optimize the hyper-parameters of an unknown function using a validation set. In particular, we use the \verb|BayesianOptimization| function from the \verb|bayesian-optimization| package implemented in Python \citep{Fernando2017-ou}.
Table~\ref{tab:parametergrid} shows the parameter ranges that are used throughout the experiments.\footnote{More information about the respective hyper-parameters can be found on the CNTK web site \emph{Python API for CNTK \url{https://cntk.ai/pythondocs/cntk.html}}} 

Furthermore, we choose input and output window sizes throughout the experiments according to the discussions in Section~\ref{sec:mva}. In particular, for the NN5 dataset, as the forecasting horizon is 56, we use an \textit{inputSize} of 70 and an \textit{outputSize} of 56. As the CIF2016 dataset has two different target horizons and some very short series, choosing window sizes is more complicated here and differs among different models, as outlined in the following.
%
%
In our experiments, we use the following variants of our proposed methodology and LSTM baseline:

 \begin{description}
\item[LSTM.Horizon] 
In the CIF2016 competition, additional knowledge is available in the form of 2 different required forecasting horizons. We use this additional knowledge to group the time series accordingly. I.e., separate prediction models are generated for each group of time horizons, following the steps 5:16 of Algorithm~\ref{alg:lstm}. Also, some of the series with a required horizon of 6 are very short and consist only of 23 data points. Following our heuristic, we use an \textit{inputSize} of 7 when the required \textit{outputSize} is 6, and an \textit{inputSize} of 15 when the required \textit{outputSize} is 12.

\item[LSTM.Cluster] Our proposed method as illustrated in Algorithm~\ref{alg:lstm}. An individual prediction model is produced for each cluster obtained. Due to the peculiarities and short series within the CIF2016 dataset, we start with the same partition as for the LSTM.Horizon model in this case, and then apply the methodology only for the series with a target horizon of 12. This is mainly because the \verb|anomalous-acm| package that we use to extract features uses internally an STL decomposition, and therefore needs 2 full periods of data, i.e., 24 data points in our case. So, as the clustering method is not applicable for some of the CIF2016 series, the LSTM.Cluster variant is only performed on 57 time series with 12 months of forecasting horizon. As a result, in addition to the LSTM models that are generated for each cluster, a separate LSTM model is built for the remaining 15 time series with forecasting horizon of 6 months.

\item[LSTM.All] The baseline LSTM algorithm, where no subgrouping is performed but one model is generated globally across all time series in the dataset. Note that, as some series from the CIF2016 dataset are very short, we use here an \textit{inputSize} of 7 throughout, and an \textit{outputSize} of 12, both for target horizons 6 and 12.

\end{description}

\subsection{Statistical tests of the results}

We perform a non-parametric aligned Friedman rank-sum test with Hochberg's post-hoc procedure to assess the presence of statistical significance of differences within multiple forecasting methods, and then to further characterize these differences~\citep{Garcia2010-jx}\footnote{More information can be found on the thematic web site of SCI2S about \emph{Statistical Inference in Computational Intelligence and Data Mining \url{http://sci2s.ugr.es/sicidm}}}. The statistical testing is done using the sMAPE measure. We also use a non-parametric paired Wilcoxon signed rank test, to examine the statistical significance of differences within two forecasting methods.

We use the statistical testing framework in two steps. First, we determine whether the differences among the proposed and the base model are statistically significant. Then, we compare our approaches to competitive benchmarks, in particular to the original participants of the competitions where possible. A significance level of $\alpha$ = 0.05 is used.

\subsection{Performed experiments}

To provide a thorough empirical study that allows us to assess accurately the performance of our approach in different situations, we perform experiments and evaluations with three different setups as follows:

 \begin{description}
 
\item[Competition setup (CO):] We run our methods under the original competitions' setup. We use a fixed origin with a withheld test set, and evaluate using the mean sMAPE, which is the primary evaluation metric used in both competitions. In this scenario, we are able to compare the proposed methods against all participants of the original competitions.
 
 
\item[Fixed origin setup (FO):] In the CIF2016 competition, the forecasts from the originally participating methods are publicly available. This enables us to perform our two-stepped statistical testing procedure and compare against all participating methods as benchmarks. 
%
From the NN5, to the best of our knowledge the actual forecasts for each method on each time series are not publicly available, and we are unable to perform testing for statistical significance on all originally participating methods. Alternatively, we use controlled benchmarks to measure the statistical significance against our approach. In particular, we use ARIMA, seasonal na\"ive, and ETS methods from the \verb|forecast| package. We also use the ES method from the \verb|smooth| package \citep{Svetunkov2017-je}, which is an alternative  implementation of exponential smoothing. Compared to its counterpart ETS from the \verb|forecast| package, ES is not restricted by the number of seasonal coefficients to be included in the model.

\item[Rolling origin setup (RO):] We also perform rolling origin evaluation using the test sets by averaging the accuracy over different forecasting origins \citep{Tashman2000-nj}. 
This enables us to obtain stronger evidence as results depend less on particular forecast origins and the evaluation is performed on more data points.
Again, as this setup differs from the original competition setup, we are unable to compare against the original participants, and evaluate against the benchmark methods. 


Following the nomenclature of \cite{Tashman2000-nj}, we use the benchmark models in a recalibration mode. Here, for each forecast origin, as iteratively new data points are included in the training set while the forecast origin moves forward, we re-fit the respective benchmark forecasting model. 
For the LSTM variants, re-fitting the entire model for each origin is costly, and we do not re-fit the model. Instead, we use an updating mode \citep{Tashman2000-nj}, where we use the initial model which is build from the original forecasting origin. New data points are used to update the model but no re-fitting of parameters is performed.


\end{description}

\subsection{Results on the CIF2016 dataset} 

\begin{table*}[!tb]
\centering
\resizebox{\columnwidth}{!}{%
\begin{tabular}{lcccccc}
	\toprule
	Method            					&Mean sMAPE 	&Median sMAPE	&Rank sMAPE 	&Mean MASE 	&Median MASE  &Rank MASE\\ \hline
	\textbf{LSTM.Cluster}				&\bf 10.53		&7.34			&10.88			&0.89		&0.59		  &10.83	\\
	\textbf{LSTM.Horizon}				&10.61			&7.92			&11.30			&0.90		&0.60		  &10.63	\\
	\textbf{LSTM.All}					&10.69			&6.72			&10.36			&0.83	&0.59		  &10.22	\\
	LSTMs and ETS				        &10.83			&6.60			&9.10			&\bf 0.79		&0.56		  &\bf 9.04		\\			
	ETS									&11.87			&6.67			&10.36			&0.84		&\bf 0.53	  &10.37	\\ 
	MLP									&12.13			&6.92			&11.01			&0.84		&0.54		  &11.13	\\		
	REST								&12.45			&7.57			&12.21			&0.90		&0.59		  &12.47	\\
	ES									&12.73			&6.51			&11.20			&0.87		&\bf 0.53	  &11.14	\\		
	FRBE								&12.90			&6.77			&10.92			&0.89		&0.57		  &10.89	\\	
	HEM									&13.04			&7.32			&12.18			&0.90		&0.59		  &12.24	\\
	Avg									&13.05			&8.02			&13.72			&0.99		&0.67		  &13.76	\\	
	BaggedETS							&13.13			&\bf 5.98		&\bf 9.08		&0.83	&0.54		  &9.63		\\		
	LSTM								&13.33			&8.20			&14.35			&0.95		&0.68		  &14.26	\\	
	Fuzzy c-regression					&13.73			&10.04			&14.74			&1.13		&0.72		  &14.73	\\		
	PB-GRNN								&14.50			&7.86			&13.46			&1.01		&0.65		  &13.61	\\		
	PB-RF								&14.50			&7.86			&13.46			&1.01		&0.65		  &13.61	\\	
	ARIMA								&14.56			&7.03			&12.55			&0.92		&0.56		  &12.51	\\			
	Theta								&14.76			&11.01			&17.49			&1.25		&0.74		  &17.63	\\		
	PB-MLP								&14.94			&8.05			&13.96			&0.99		&0.68		  &13.93	\\		
	TSFIS								&15.11			&10.18			&16.85			&1.27		&0.91		  &16.82	\\		
	Boot.EXPOS							&15.25			&6.92			&12.44			&0.93		&0.61		  &12.38	\\		
	MTSFA								&16.51			&9.69			&15.51			&1.13		&0.71		  &15.46	\\		
	FCDNN								&16.62			&8.71			&17.21			&1.14		&0.82		  &17.35	\\	
	Na{\"i}ve Seasonal					&19.05			&12.72			&19.68			&1.29		&0.95		  &20.02	\\	
	MSAKAF								&20.39			&14.24			&18.92			&1.57		&1.31		  &18.86	\\		
	HFM									&22.39			&11.89			&19.69			&3.27		&1.14		  &19.84	\\	
	CORN          						&28.76			&19.86			&24.68			&2.24		&1.83		  &24.66	\\ \hline
\end{tabular}%
}
\caption{Results for the 72 monthly series of CIF2016, ordered by the first column, which is the Mean sMAPE. For each column, the results of the best performing method(s) are marked in boldface.}

\label{tab:cif}
\end{table*}


\begin{table*}[!tb]
\centering
\footnotesize
\begin{tabular}{llrrr}
	\toprule
	Method			&$p_{Hoch}$&\\ \hline
	BaggedETS		&-\\
	LSTMs and ETS	&0.634\\
	ES				&0.634\\	
	ETS				&0.634\\	
	MLP				&0.634\\
	FRBE			&0.273\\
	Boot.EXPOS		&0.273\\
	HEM				&0.216\\
	\textbf{LSTM.ALL} &0.216\\
	REST			&0.216\\
	ARIMA			&0.154\\
	\textbf{LSTM.Cluster} &0.092\\
	PB.RF				  &0.056\\ 	
	PB.GRNN				  &0.056\\
	\hline		
	\textbf{LSTM.Horizon} &0.048\\
	PB.MLP				  &0.042\\
	MTSFA				  &2.22 $\times$ $10^-4$ \\ 
	Avg					  &1.45 $\times$ $10^-4$ \\
	Fuzzy.c.regression	  &1.43 $\times$ $10^-4$ \\
	LSTM				  &7.55 $\times$ $10^-5$ \\
	TSFIS				  &3.27 $\times$ $10^-6$ \\ 
	FCDNN				  &1.31 $\times$ $10^-4$ \\
	Theta				  &1.28 $\times$ $10^-9$ \\ 
	MSAKAF				  &5.77 $\times$ $10^-11$ \\ 
	HFM					  &5.68 $\times$ $10^-12$ \\ 
	Na{\"i}ve Seasonal	  &3.34 $\times$ $10^-12$ \\ 
	CORN				  &4.51 $\times$ $10^-23$ \\
	\hline			
\end{tabular}
\caption{Results of statistical testing for the CIF2016 dataset, including original participants' results and our results (printed in boldface). Adjusted p-values calculated from the aligned Friedman test with Hochberg’s post-hoc procedure are shown. A horizontal line is used to separate the methods that perform significantly worse than the best method from the ones that do not. We see that the LSTM.All and the proposed LSTM.Cluster variants do not perform significantly worse compared to the best method BaggedETS.}
\label{tab:cifallstat}
\end{table*}






Table~\ref{tab:cif} shows the results of the CO and FO evaluation setups for the CIF2016 dataset. We see that regarding the mean sMAPE, which is the primary measure used in the original competition, the LSTM.Cluster variant from our proposed methods outperforms all other methods from the competition. In particular, it also outperforms the baseline LSTM.All variant, and all LSTM variants outperform the ETS, BaggedETS, and Theta methods, which can be seen as the state of the art for forecasting monthly data.
Considering the other measures, the results change quite drastically, and traditional univariate methods such as BaggedETS often outperform our methods. This can be attributed to the peculiarities of the CIF2016 dataset. The dataset contains 48 artificially generated time series that may not contain useful cross-series information. This claim is strengthened by the fact that already the BaggedETS submission to the competition outperformed all other methods in this subset, in particular the LSTM methods \citep{Stepnicka2016-bu}. 

For the first step of our statistical testing evaluation, where we compare the LSTM variants among themselves, the corresponding aligned Friedman test has an overall $p$-value of 0.8010, which highlights that these differences are insignificant and all LSTM methods should be considered to have comparable performance on this dataset. 

As the second step of our statistical testing evaluation, we compare the LSTM variants against the other participant methods from the CIF2016 competition. The aligned Friedman test for multiple comparisons results in an overall $p$-value of 7.84 $\times$ $10^-6$. Hence, these differences are highly significant, and Table~\ref{tab:cifallstat} shows the associated post-hoc testing. The BaggedETS method achieves the best ranking and is used as the control method. We see that the LSTM.All and LSTM.Cluster variants do not perform significantly worse than this control method.

\begin{table*}[!tb]
\centering
\resizebox{\columnwidth}{!}{%
\begin{tabular}{lcccccc}
	\toprule
	 Method    				&Mean sMAPE  &Median sMAPE 	&Rank sMAPE &Mean MASE  &Median MASE	&Rank MASE	\\ \hline
	\textbf{LSTM.Horizon}	&\bf 9.80	&7.08			&5.50		&0.79		&0.61				&5.38	\\
	\textbf{LSTM.Cluster}	&9.95		&7.01			&3.75		&0.81		&0.61				&3.88	\\
	 ETS		   			&10.76		&6.59			&3.25		&0.78		&0.52				&3.06	\\	
	 BaggedETS			   	&10.77		&\bf 5.97		&\bf 2.12	&\bf 0.76	&\bf 0.48			&\bf 2.31	\\
	 ES						&11.04		&6.56			&3.00		&0.78		&0.49				&3.06	\\
	 \textbf{LSTM.All}		&11.10		&6.97			&4.25		&0.78		&0.63				&4.25	\\
	 ARIMA					&12.76		&7.28			&6.38		&0.83		&0.55				&6.31	\\
	 Na{\"i}ve Seasonal		&19.82		&12.09			&7.75		&1.38		&0.99				&7.75	\\
	 \hline              
\end{tabular}
}
\caption{Results of the rolling origin evaluation on selected benchmarks, using the Mean sMAPE measure, for the 72 monthly series of CIF2016, in ascending order.  For each column, the results of the best performing method(s) are marked in boldface.}
\label{tab:rollingorigin}
\end{table*}

Table~\ref{tab:rollingorigin} shows the evaluation summary of the RO setup, where the proposed LSTM.Horizon variant obtains the best Mean sMAPE, outperforming the baseline LSTM.All variant and all other state-of-the-art univariate forecasting methods such as ETS, BaggedETS, ES, ARIMA, and Na{\"i}ve Seasonal. On all the other error measures, BaggedETS performs best, which is consistent with the particularities of this dataset as discussed earlier.


%



\subsection{Results on the NN5 dataset} 

\begin{table*}[!tb]
\centering
\footnotesize
\begin{tabular}{lrrrr}
	\toprule
	 Contender Name    &Mean sMAPE   \\ \hline
	 Wildi			   &19.9		 \\
	 Andrawis		   &20.4		 \\	
	 Vogel			   &20.5		 \\
	 D'yakonov		   &20.6		\\
	 Noncheva		   &21.1		 \\
	 \textbf{LSTM.Cluster}	   &\textbf{21.6}\\
	 Rauch		   	   &21.7		 \\
	 Luna			   &21.8		 \\
	 Lagoo			   &21.9		\\
	 Wichard		   &22.1		\\
	 Gao		       &22.3		\\
	 \textbf{LSTM.All}	   &\textbf{23.4}\\
	 Puma-Villanueva   &23.7	     \\
	 Autobox(Reilly)   &24.1		 \\
	 Lewicke    	   &24.5		  \\
	 Brentnall   	   &24.8		 \\
	 Dang   		   &25.3	      \\ 
	 Pasero			   &25.3	     \\
	 Adeodato		   &25.3	     \\
	 undisclosed	   &26.8	      \\
	 undisclosed	   &27.3	     \\
	 Tung   		   &28.1	     \\
	 Na{\"i}ve Seasonal&28.8		 \\
	 undisclosed	   &33.1	      \\
	 undisclosed	   &36.3	     \\
	 undisclosed	   &41.3	      \\
	 undisclosed	   &45.4	     \\
	 Na{\"i}ve Level	   &48.4	\\
	 undisclosed  		   &53.5 \\
	 \hline              
\end{tabular}
\caption{Original Mean sMAPE results for the 111 daily series of the NN5, together with the results from our methods, in ascending order. 
}
\label{tab:nn5}
\end{table*}

\begin{table*}[!tb]
\centering
\resizebox{\columnwidth}{!}{%
\begin{tabular}{lcccccc}
	\toprule
	 Method            		&Mean sMAPE 	&Median sMAPE  &Rank sMAPE  &Mean MASE  &Median MASE  &Rank MASE  \\ \hline
	 ES          			&\bf 21.44		&\bf 20.29	   &3.12        &\bf 0.86		&\bf 0.80		  &\bf 2.91	\\
	 ETS					&21.46			&20.57		   &\bf 3.07		&\bf 0.86		&0.81	      &3.05	\\
	 BaggedETS				&21.46			&20.57		   &\bf 3.07	    &0.87       &0.84		  &3.19	\\								
	\textbf{LSTM.Cluster}	&21.66			&20.71		   &3.85	    &0.94		&0.89         &4.35\\		
	\textbf{LSTM.All}		&23.46			&21.75		   &4.51		&0.96		&0.93         &5.04 \\	
	ARIMA					&25.29			&21.74		   &4.66		&0.97		&0.90		  &4.32\\	
	 Na{\"i}ve Seasonal		&26.49			&23.31		   &5.73		&1.01		&0.94	      &5.15\\	
	 \hline
\end{tabular}
}
\caption{Results of the fixed origin evaluation for the 111 daily series of NN5, ordered by the first column, which is the Mean sMAPE.  For each column, the results of the best performing method(s) are marked in boldface.}
\label{tab:nn5_add}
\end{table*}

\begin{table*}[!tb]
\centering
\resizebox{\columnwidth}{!}{%
\begin{tabular}{lcccccc}
	\toprule
	 Method    				&Mean sMAPE &Median sMAPE 	&Rank sMAPE &Mean MASE &Median MASE &Rank MASE\\ \hline
	 ES						&\bf 22.18	&20.75			&2.66		&\bf 0.87	&\bf 0.83   &2.50   \\
	 ETS		   			&22.24		&\bf 20.56		&\bf 2.63	&\bf 0.87	&\bf 0.83	&\bf 2.47	\\
	 BaggedETS				&22.39		&20.80			&3.00		&0.88		&0.86		&3.00	 \\
	 \textbf{LSTM.Cluster}	&23.38		&22.27			&4.18		&0.93		&0.89		&4.30	\\
	 \textbf{LSTM.All}		&23.89		&22.44			&4.87		&0.95		&0.93		&4.83	\\
	 ARIMA					&23.96		&22.61			&3.85		&0.96		&0.94		&4.11	\\
	 Na{\"i}ve Seasonal		&28.23		&26.10			&6.81		&1.10		&1.04       &6.78 \\
	 \hline              
\end{tabular}
}
\caption{Results of the rolling origin evaluation, ordered by the Mean sMAPE measure, for the 111 daily series of NN5, in ascending order.  For each column, the results of the best performing method(s) are marked in boldface.}
\label{tab:nn5rollingorigin}
\end{table*}


\begin{table*}[!tb]
\centering
\footnotesize
\begin{tabular}{llrrr}
	\toprule
	Method					&$p_{Hoch}$&\\ \hline
	ETS						&-\\
	BaggedETS				&1\\
	ES						&1\\
	\textbf{LSTM.Cluster}	&0.233\\
	\hline
	\textbf{LSTM.All}		&5.03 $\times$ $10^-8$\\
	 ARIMA					&4.28 $\times$ $10^-11$\\
	 Na{\"i}ve Seasonal		&3.77 $\times$ $10^-22$\\
	\hline			
\end{tabular}
\caption{Results of statistical testing for NN5 data, using the results of the selected benchmarks and the LSTM variants. ETS performs best, and ES, BaggedETS, and LSTM.Cluster do not perform significantly worse.}
\label{tab:nn5allstat}
\end{table*}

Table~\ref{tab:nn5} shows the original results of the CO setup for the NN5 forecasting competition data, together with results of our methods. 
It can be seen that the proposed LSTM.Cluster variant performs better than the LSTM.All variant, and reaches a 6\textsuperscript{th} overall rank. 
Note that the proposed LSTM.Horizon variant is not benchmarked against the NN5 dataset, as no additional information is available in this case. 

Table~\ref{tab:nn5_add} and Table~\ref{tab:nn5rollingorigin} provide results of the FO and RO setup evaluations on the NN5 dataset. 
As for the NN5 the forecasts from the original participants are not available, both FO and RO evaluations, as well as the statistical testing, are performed using the benchmark methods.

We see that the LSTM.Cluster variant consistently outperforms the LSTM.ALL variant across all error measures and both for fixed and rolling origin. However, the methods are not able to outperform the exponential smoothing benchmarks, ES and ETS.  



The paired Wilcoxon signed-rank test gives an overall $p$-value of 7.22 $\times$ $10^-4$, within the two LSTM variants. Therefore, the differences among the LSTM variants, LSTM.Cluster and LSTM.All are highly significant.

Table~\ref{tab:nn5allstat} shows the results of the second step of our statistical testing evaluation. The overall result of the aligned Friedman rank sum test is a $p$-value of 6.59 $\times$ $10^-11$, which is highly significant. The ETS method performs best and is used as the control method. Also, according to Table~\ref{tab:nn5allstat}, we see that the baseline LSTM method performs significantly worse than the control method, whereas the LSTM.Cluster variant does not.

\section{Conclusions}
\label{sec:conc}

Nowadays, large quantities of related and similar time series are available in many application cases.
To exploit the similarities between multiple time series, recently methods to build global models across such time series databases have been introduced. One very promising approach in this space are Long Short-Term Memory networks, a special type of recurrent neural networks.

However, in the presence of disparate time series, the accuracy of such a model may degenerate, and accounting for the notion of similarity between the time series becomes necessary. Motivated by this need, we have proposed a forecasting framework that exploits the cross-series information in a set of time series by building separate models for subgroups of time series, specified by an automatic clustering methodology.

We have evaluated our proposed methodology on two benchmark competition datasets, and have achieved competitive results. On the CIF2016 dataset, our methods outperform all the other methods from the competition with respect to the evaluation metric used in the competition, and in the NN5 competition dataset our method ranks 6\textsuperscript{th} overall, and achieves consistent improvements over the baseline LSTM model.
The results indicate that the LSTM is a competitive method, effectively exploiting similarities of the time series and therewith being able to outperform state-of-the-art univariate forecasting methods. Subgrouping of similar time series with our proposed methodology augments the accuracy of this baseline LSTM model in many situations.

\section*{References}

\bibliographystyle{elsarticle-harv}
\bibliography{reference}

\end{document}